\documentclass[conf]{new-aiaa}
\usepackage[utf8]{inputenc}

\usepackage{graphicx}
\usepackage{amsmath}
\usepackage[version=4]{mhchem}
\usepackage{siunitx}
\usepackage{longtable,tabularx}
\usepackage{dsfont}
\usepackage{float}
\usepackage{array}
\newcolumntype{C}[1]{>{\centering\arraybackslash}p{#1}}

\newcommand{\vect}[1]{\boldsymbol{#1}}
\newcommand{\atanTwo}{\operatorname{atan2}}
\newcommand{\mylist}[1]{\overline{\MakeUppercase{#1}}}

\newcommand\blfootnote[1]{
    \begingroup
    \renewcommand\thefootnote{}\footnote{#1}
    \addtocounter{footnote}{-1}
    \endgroup
}

\title{Probabilistic Weapon Engagement Zones \\for a Turn Constrained Pursuer}
\author{Grant Stagg\footnote{PhD Student, Electrical and Computer Engineering, Brigham Young University.}}
\author{Cameron K. Peterson\footnote{Associate Professor, Electrical and Computer Engineering, Brigham Young University, and AIAA Senior Member.}}
\affil{Brigham Young University, Provo, Utah, 84602, USA}
\author{Isaac E. Weintraub\footnote{Senior Engineer, Control Science Center, Air Force Research Laboratory, Wright-Patterson AFB OH 45433, AIAA Lifetime Associate Fellow}}
\affil{Air Force Research Laboratory, Wright-Patterson AFB, Ohio, 45433, USA}

\begin{document}
\maketitle
\blfootnote{Distribution Statement A. Approved for Public Release: Distribution is Unlimited (AFRL-2025-5483, Cleared 05 DEC 2025)}

\begin{abstract}
Curve–straight probabilistic engagement zones (CSPEZ) quantify the spatial regions an evader should avoid to reduce capture risk from a turn-rate-limited pursuer following a curve–straight path with uncertain parameters (position, heading, velocity, range, and maximum turn rate). This paper presents methods for generating evader trajectories that minimize capture risk under such uncertainty. An analytic solution is first derived for the deterministic curve–straight basic engagement zone (CSBEZ), which is then extended to a probabilistic framework through sensitivity analysis using Monte Carlo sampling, linearization, quadratic approximation, and neural network regression. The accuracy and computational cost of these approaches are evaluated in a simulation. Finally, CSPEZ constraints are incorporated into a trajectory optimization algorithm to produce safe paths that explicitly account for pursuer uncertainty.
\end{abstract}


\section{Introduction}
\lettrine{N}avigating hostile or contested areas is a common requirement in many mission scenarios, particularly for unmanned aerial vehicle (UAV) reconnaissance. These vehicles must often operate in environments where detection or engagement by adversaries presents a significant risk to the platform. To ensure mission success, it is essential to generate safe and efficient trajectories~\cite{jun2003path,peng2012intelligent} that minimize the need for abrupt evasive actions during execution~\cite{von2023basic}. A key challenge arises when knowledge of the adversarial environment—such as the positions or capabilities of threats—is incomplete or uncertain. If not properly accounted for, this uncertainty can lead to an over- or underestimation of the path risk. Therefore, robust planning approaches must explicitly model uncertainty to ensure viable and survivable routes.

In this work, we use engagement zones (EZ) to model adversarial threats. EZs are regions an evader must avoid to guarantee non-engagement with a pursuer. Past work has defined EZs using cardioid shapes~\cite{weintraub2022optimal,dillon2023optimal,wolek2024sampling}. These works created path-planning algorithms for an evader to avoid engagements~\cite{weintraub2022optimal}, for accounting for two EZs~\cite{dillon2023optimal}, and for many EZs~\cite{wolek2024sampling}. All these methods use cardioid-shaped EZs that do not directly represent the pursuer's capabilities. These formulations also assume perfect knowledge of the pursuer’s parameters and do not account for pursuer or environmental uncertainty.

Unlike cardioid-shaped EZs, basic engagement zones (BEZs)~\cite{von2023basic} are derived from the principles of differential games~\cite{isaacs1965differential} and define the region in which a pursuer can capture a target. Assuming an infinite turn rate, the BEZ model was applied to generate safe trajectories around a single pursuer. Building on this, in a previous work~\cite{stagg2025}, we extended the BEZ framework to account for uncertainty in the evader and pursuer parameters through probabilistic engagement zones (PEZs). We used a linear approximation to propagate uncertainty through the BEZ equation and developed a path-planning algorithm that incorporated the linear PEZ approximation as a constraint. However, the infinite–turn–rate assumption is overly conservative for real-world pursuers, and in this work, we provide a less conservative model that incorporates turn-rate-constrained vehicles.

In recent work, the BEZ was extended to use a Dubin's vehicle model~\cite{dubins1957curves} for the pursuer~\cite{chapman2025engagement}. Two BEZ variants were introduced to better represent the pursuer’s maneuvering capabilities: the curve-only BEZ (CBEZ), in which the pursuer follows a single arc of constant curvature to intercept the evader, and the curve-straight BEZ (CSBEZ), in which the pursuer executes a minimum-radius turn (i.e., at maximum turn rate) followed by a straight segment to complete the interception. Their work derived an analytic solution for the CBEZ and a semi-analytic solution for the CSBEZ, assuming known pursuer parameters. Both formulations require precise knowledge of pursuer parameters: position, heading, maximum turn-rate, range, and speed.  This work extends that formulation to account for uncertainty in pursuer parameters

Parallel work in addressing path planning under uncertainty in adversarial settings has explored the use of stochastic differential games~\cite{friedman1972stochastic,patil2023risk,li2007two}. In particular,~\cite{patil2023risk} applies path integral control, a sampling-based technique, to solve zero-sum stochastic differential games. This sampling-based approach is computationally expensive for real-time use. The researchers in~\cite{li2007two} use optimal control to solve the stochastic differential game for the turn-rate limited pursuer and evader. These methods are primarily reactive, offering strategies for agents after the engagement has begun. These methods optimize solely for capture avoidance and do not account for higher-level mission objectives. In contrast, the framework presented here provides constraints that can integrate with higher-level mission objectives to ensure safety while accounting for uncertainty. 

In this paper, we extend the prior research on CSBEZ to provide safer trajectory planning for UAVs operating in adversarial environments.  Our contributions include:
an analytic solution for the CSBEZ that was originally introduced in~\cite{chapman2025engagement}; an extension of the CSBEZ to account for uncertainty; and a path optimization algorithm that accounts for this uncertainty when generating low-risk trajectories. Specifically, we model uncertainty in the pursuer's position, heading, turn rate, range, and velocity through the introduction of the curve-straight \emph{probabilistic} engagement zone (CSPEZ). We propose four methods for solving the CSPEZ. Two are based on local approximations: linear CSPEZ (LCSPEZ) and quadratic CSPEZ (QCSPEZ). One is a Monte Carlo baseline (MCCSPEZ) used for comparison and to approximate ground truth. Finally, the neural-network-based NNCSPEZ employs a multilayer perceptron trained on simulated data. We provide a numerical comparison of all these methods and describe the trade-offs in using each one. A trajectory optimization algorithm is also developed that incorporates CSPEZ constraints to ensure path safety under parameter uncertainty.

The remainder of the paper is organized as follows. Section~\ref{sec:CSBEZ} presents the background and definition of the CSBEZ, including an analytical derivation. Section~\ref{sec:CSPEZ} extends this model to incorporate uncertainty in the pursuer’s parameters, introducing the curve–straight probabilistic engagement zone (CSPEZ) and describing four approximation methods. Section~\ref{sec:path} details path-planning algorithms that leverage the CSPEZ formulations. Section~\ref{sec:results} presents the experimental results, and Section~\ref{sec:conclusion} concludes the paper.

\section{Curve-Straight Basic Engagement Zone (CSBEZ)}\label{sec:CSBEZ}

The curve–straight basic engagement zone (CSBEZ) defines the configurations (spatial regions and headings) an evader must avoid to prevent capture by a turn-rate–constrained pursuer executing a curve–straight path. It is defined in terms of the pursuer's parameters:  initial position $P$, speed $v_P$, heading $\psi_P$, range $R$ (equivalent to maximum flight time $R = v_Pt_{\text{max}}$), minimum turn radius $a$, as well as the evader's initial position $E$, speed $v_E$, and heading $\psi_E$. To determine whether an evader lies within the CSBEZ, we test whether the pursuer can intercept the evader by traveling a distance less than $R$, assuming the evader maintains its current heading.

The geometry of the engagement zone is obtained by considering the evader's projected path.  If the evader maintains its current heading for the length of the pursuer's maximum flight time, the evader will arrive at 
\begin{equation}
F = E + \nu R \vect{\hat{v}}_E,
\end{equation}
where $\nu = v_E/v_P$, $\vect{\hat{v}}_E\in \mathbb{R}^2$ is a unit vector that points in direction $\psi_E$, $||\vect{\hat{v}}_E||_2=1$ and $\psi_E = \angle \vect{\hat{v}}_E = \mathrm{atan2}(\vect{\hat{v}}_E[2],\vect{\hat{v}}_E[1] )$.  We assume the pursuer takes a curve-straight (CS) path to intercept the evader at $F$. This consists of a single arc (left or right) of radius of $a$ and angle $\theta$,  starting from the pursuer's initial position $P$ and ending at a point $G$, followed by a straight-line segment from $G$ to $F$. The line segment $\overline{GF}$ is tangent to the arc. The total path length is $\theta a + \|F-G\|_2$. If this distance is less than $R$ the evader is inside the EZ. 

To compute the path length, we must first determine the point $G$, where the pursuer transitions from the turning arc to the straight line segment toward $F$. The point $G$ lies at the intersection where line segment $\overline{GF}$ falls along the tangent line to the turn circle, which can be determined geometrically from the turn-center location. To compute these points, we first consider the point from the left-turn circle, denoted as $G_{\ell}$.  The center of the left-turn circle is
\begin{equation}
    C_{\ell}(P,\psi_P,a) = P + a \begin{bmatrix}
        -\sin(\psi_P)\\
        \cos(\psi_P)
    \end{bmatrix}.
\end{equation}

\begin{figure}[htbp]
    \centering
\includegraphics[width=1.0\linewidth,clip]{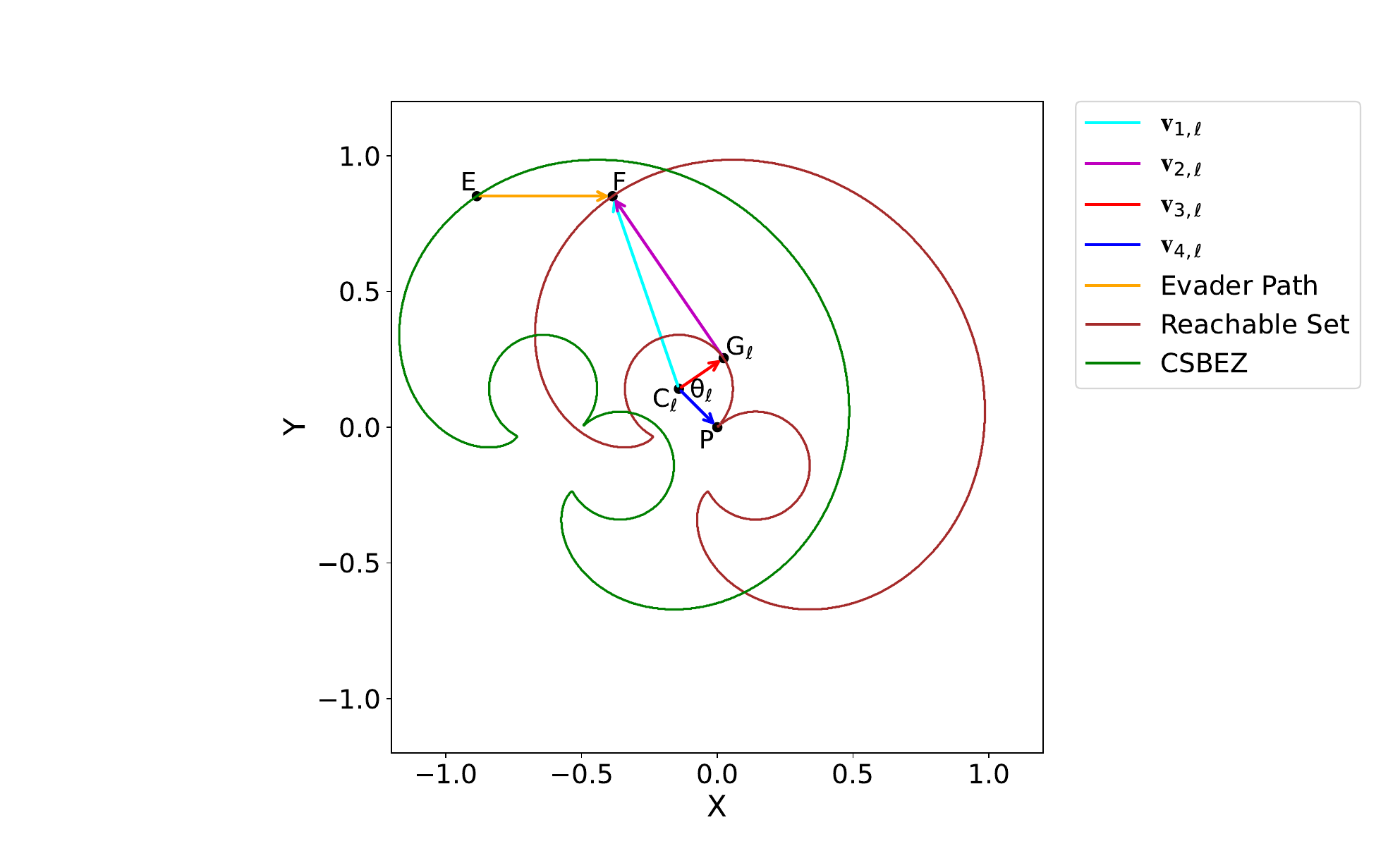}
\caption{Reachable set (brown) and CSBEZ (green). The evader’s start position $E$ and projected position $F$ are shown, along with the pursuer’s start $P$, point $G_\ell$, and turn center $C_\ell$. The four vectors $\vect{v}_{1,\ell}$–$\vect{v}_{4,\ell}$ used in the path-length derivation are also illustrated.}\label{fig:CSBEZ}
\end{figure}

To aid in finding $G_\ell$ we define four vectors that are functions of the pursuer's parameters ($P, \psi_P,$ and $a$) and $F$:
\begin{subequations}\label{eq:vectors}
\begin{align}
\vect{v}_{1,\ell}(F,P,\psi_P,a) &= F - C_{\ell}(P,\psi_P,a) \label{eq:v1}\\
\vect{v}_{2,\ell}(F,P,\psi_P,a) &= F - G_{\ell}(F,P,\psi_P,a) \label{eq:v2}\\
\vect{v}_{3,\ell}(F,P,\psi_P,a) &= G_{\ell}(F,P,\psi_P,a) - C_{\ell}(P,\psi_P,a) \label{eq:v3}\\
\vect{v}_{4,\ell}(F,P,\psi_P,a) &= P - C_{\ell}(P,\psi_P,a) \label{eq:v4}
\end{align}
\end{subequations}
These vectors correspond to: (a) the line from the left-turn center to the intercept point, (b) the tangent from the turn to the intercept point, (c) the radius from the left-turn center to the point $G$, and (d) the line from the left-turn center to the pursuer’s initial position. These relationships are illustrated in Figure~\ref{fig:CSBEZ}, along with the EZ and pursuer's reachable set.  Next, we show how these vectors may be derived.

For clarity, we omit the functional dependence of the vectors, though they remain functions of the pursuer’s parameters. Of these vectors, two are known ($\vect{v}_{1,\ell}$ and $\vect{v}_{4,\ell}$) and two depend on the point $G_\ell$ ($\vect{v}_{2,\ell}$ and $\vect{v}_{3,\ell}$). The vector $\vect{v}_{1,\ell}$ points from the left-turn center to the target point $F$. To solve for $G_{\ell}$, we solve for $\vect{v}_{3,\ell}$, the vector from $C_\ell$ to $G_\ell$. We know that the straight path the pursuer takes after it stops turning ($\vect{v}_{2,\ell}$) is tangent to the circle. Therefore, $\vect{v}_{2,\ell}$ and $\vect{v}_{3,\ell}$ are perpendicular:
\begin{equation}
    \vect{v}_{2,\ell} \cdot \vect{v}_{3,\ell} = 0.
\end{equation}
We also note that $\vect{v}_{1,\ell}$ is a linear combination of $\vect{v}_{2,\ell}$ and $\vect{v}_{3,\ell}$, so
\begin{equation}
    \vect{v}_{2,\ell} = \vect{v}_{1,\ell} - \vect{v}_{3,\ell}. 
\end{equation}
Substituting this into the orthogonality condition, we obtain a dot product expression involving 
$\vect{v}_{1,\ell}$ and $\vect{v}_{3,\ell}$. Since 
$\vect{v}_{3,\ell}$ lies on a circle of radius $a$, we can enforce the constraint 
$\|\vect{v}_{3,\ell}\| = a$. This yields the following relation:
\begin{equation}\label{eq:v1_v3_dot}
\vect{v}_{1,\ell} \cdot \vect{v}_{3,\ell} = a^2.
\end{equation}
The vector $\vect{v}_{3,\ell}$ can be decomposed into components parallel and perpendicular to $\vect{v}_{1,\ell}$; the parallel component contributes to the dot product, while the perpendicular component does not.  The perpendicular component can be used to ensure $\|\vect{v}_{3,\ell}\|_2 = a$.  This decomposition takes the form
\begin{equation}
\vect{v}_{3,\ell} = \alpha \vect{v}_{1,\ell} + \beta \vect{v}_{1,\ell}^\perp,
\end{equation}
where $\vect{v}_{1,\ell}^\perp$ is a unit vector perpendicular to $\vect{v}_{1,\ell}$.
Two perpendicular directions exist, one formed by a \ang{90} clockwise (CW) rotation and the other by a counter-clockwise (CCW) rotation. To ensure $\vect{v}_{3,\ell}$ is pointing the correct direction we choose the CW perpendicular vector for $\vect{v}_{1,\ell}^\perp$:
\begin{equation}
    \vect{v}_{1,\ell}^\perp = \begin{bmatrix}
        0 & 1 \\ -1 & 0
    \end{bmatrix}\frac{\vect{v}_{1,\ell}}{\|\vect{v}_{1,\ell}\|}.
\end{equation}
To solve for $\alpha$ we use the dot product relationship defined in Equation~\eqref{eq:v1_v3_dot}. Since we decomposed $\vect{v}_{3,\ell}$ into a parallel and orthogonal components, only the parallel component contributes to the dot product:
\begin{equation}
\vect{v}_{1,\ell} \cdot \vect{v}_{3,\ell} =\alpha \vect{v}_{1,\ell} \cdot \vect{v}_{1,\ell} + \beta \vect{v}_{1,\ell} \cdot \vect{v}_{1,\ell}^\perp= \alpha \|\vect{v}_{1,\ell}\|_2^2=a^2
\end{equation}
which yields
\begin{equation}\label{eq:alpha}
\alpha = \frac{a^2}{\|\vect{v}_{1,\ell}\|_2^2}.
\end{equation}
Next, we use the constraint $\|\vect{v}_{3,\ell}\|_2 = a$ to solve for the perpendicular component of $\vect{v}_{3,\ell}$. 
 Squaring both sides gives:
\begin{equation}\label{eq:norm_squared_v3}
a^2 = \vect{v}_{3,\ell} \cdot \vect{v}_{3,\ell}.
\end{equation}
Substituting the decomposition 
$\vect{v}_{3,\ell} = \alpha \vect{v}_{1,\ell} + \beta \vect{v}_{1,\ell}^\perp$, 
we expand the dot product:
\begin{align}
a^2 &= (\alpha \vect{v}_{1,\ell} + \beta \vect{v}_{1,\ell}^\perp) \cdot (\alpha \vect{v}_{1,\ell} + \beta \vect{v}_{1,\ell}^\perp) \\
&= \alpha^2 \vect{v}_{1,\ell} \cdot \vect{v}_{1,\ell} 
+ 2\alpha \beta \vect{v}_{1,\ell} \cdot \vect{v}_{1,\ell}^\perp 
+ \beta^2 \|\vect{v}_{1,\ell}^\perp\|^2
\end{align}
We now simplify the expression. Since $\vect{v}_{1,\ell}$ and $\vect{v}_{1,\ell}^\perp$ are orthogonal, the cross term vanishes, $\vect{v}_{1,\ell} \cdot \vect{v}_{1,\ell}^\perp = 0$
and because $\vect{v}_{1,\ell}^\perp$ is a unit vector, we have $\|\vect{v}_{1,\ell}^\perp\|^2 = 1$. Equation~\eqref{eq:norm_squared_v3} simplifies to:
\begin{equation}
a^2 = \alpha^2 \|\vect{v}_{1,\ell}\|^2 + \beta^2
\end{equation}
Substituting in the value of $\alpha = \dfrac{a^2}{\|\vect{v}_{1,\ell}\|^2}$ (Equation~\eqref{eq:alpha}), we get:
\begin{equation}
a^2 = \frac{a^4}{\|\vect{v}_{1,\ell}\|^2} + \beta^2.
\end{equation}
Solving for $\beta$ yields:
\begin{equation}
\beta = \sqrt{a^2 - \frac{a^4}{\|\vect{v}_{1,\ell}\|^2}}.
\end{equation}
Using the parallel ($\alpha$) and perpendicular ($\beta$) components $\vect{v}_{3,\ell}$ is 
\begin{equation}
\vect{v}_{3,\ell}(F,P,\psi_P,a) = \frac{a^2}{\|\vect{v}_{1,\ell}\|^2} \vect{v}_{1,\ell}+\sqrt{a^2 - \frac{a^4}{\|\vect{v}_{1,\ell}\|^2}} 
\begin{bmatrix} 0 & 1 \\ -1 & 0 \end{bmatrix} 
\frac{\vect{v}_{1,\ell}}{\|\vect{v}_{1,\ell}\|} .
\end{equation}
Given $\vect{v}_{3,\ell}$, the vector from the left-turn center $C_\ell$ to the point $G_\ell$, we can now compute the point as
\begin{equation}
G_\ell(F,P,\psi_P,a) = C_{\ell}+\vect{v}_{3,\ell}(F,P,\psi_P,a).
\end{equation}
The point $G_\ell$ can be used to compute the total length of the pursuer’s curve-straight path to the evader’s projected position $F$. This path consists of two segments: an arc from the pursuer’s initial position $P$ to the point $G_\ell$, followed by a straight-line segment from $G_\ell$ to $F$.
To compute the arc length, we first determine the turning angle swept by the pursuer.
This angle is measured from the vector pointing from the center of the left-turn circle to the pursuer’s initial position, $\vect{v}_{4,\ell}$, to the vector pointing from the center to the point, $\vect{v}_{3,\ell}$. Because the turn is CCW, the turning angle is computed from the two-dimensional cross and dot products using the $\atanTwo$ function:
\begin{equation}
    \theta_{ccw}(F, P, \psi_P, a) = \atanTwo(\vect{v}_{4,\ell}\times\vect{v}_{3,\ell},\vect{v}_{4,\ell} \cdot \vect{v}_{3,\ell}).
\end{equation}
The total path length is the sum of the arc length and the straight-line distance from the $G$ to $F$
\begin{equation}
    L_{\ell}(F,P, \psi_P,a) = a \theta_{ccw}(F, P, \psi_P, a) + \|G_{\ell}(F,P,\psi_P,a) - F \|_2.
\end{equation}
If we need to instead compute the shortest right-straight path, then the only difference is which perpendicular vector we choose when constructing $\vect{v}_{3,r}$.  We compute the CW angle instead of the CCW angle, and choose the right center point. In the right-straight case 
\begin{equation}
    C_r(P,\psi_P,a) = P + a \begin{bmatrix}
        \sin(\psi_P)\\
        -\cos(\psi_P)
    \end{bmatrix},
\end{equation}
\begin{equation}
\vect{v}_{3,r} = \frac{a^2}{\|\vect{v}_{1,r}\|^2} \vect{v}_{1,r}+\sqrt{a^2 - \frac{a^4}{\|\vect{v}_{1,r}\|^2}} 
\begin{bmatrix} 0 & -1 \\ 1 & 0 \end{bmatrix} 
\frac{\vect{v}_{1,r}}{\|\vect{v}_{1,r}\|},
\end{equation}
and 
\begin{equation}
    \theta_{cw}(F, P, \psi_P, a) = \atanTwo{(-\vect{v}_{4,r}}\times\vect{v}_{3,r},\vect{v}_{4,r} \cdot \vect{v}_{3,r}).
\end{equation}
Using this angle, we define the shortest right-turn length as
\begin{equation}
    L_r(F,P, \psi_P,a) = a \theta_{cw}(F, P, \psi_P, a) + \|G_r(F,P,\psi_P,a) - F \|_2.
\end{equation}
The overall shortest curve-straight length is 
\begin{equation}
    L(F,P, \psi_P,a) = \min(L_r(F, P, \psi_P, a),L_{\ell}(F, P, \psi_P, a)).
\end{equation}
If $F$ lies to the right of the Pursuer's heading, then the right-straight path will be shorter than the left-straight path.
We define the EZ function as
\begin{equation}\label{eq:ez_equation}
    z(E,\psi_E,P,\psi_P,a,R,\nu) = L(E+\nu R \vect{\hat{v}}_{E},P,\psi_P,a) - R.
\end{equation}
The CSBEZ comprises all the points for which the minimum curve-straight path length required to reach the evader's future position, after the pursuer has traveled its maximum range, is less than $R$.  The CSBEZ is defined as
\begin{equation}
    \mathcal{Z} = \{E | z(E,\psi_E,P,\psi_P,a,R,\nu) \leq 0 \}.
\end{equation}
Figure~\ref{fig:CSBEZ} shows both the reachable set and the CSBEZ. Also shown are the points and vectors that were used to compute the left-straight path length.

\section{Curve-Straight Probabilistic Engagement Zone (CSPEZ)}\label{sec:CSPEZ}
The previous section derived the CSBEZ assuming the pursuer's parameters are known. In adversarial environments, however, the pursuer's parameters are often unknown or uncertain. The CSPEZ quantifies the probability that the evader lies within the true CSBEZ when uncertainty in the pursuer's parameters is considered. This probability is approximated using four uncertainty-propagation techniques, described in the following sections:
Monte Carlo CSPEZ (MCCSPEZ), linearized CSPEZ (LCSPEZ), quadratic CSPEZ (QCSPEZ), and neural network CSPEZ (NNCSPEZ).

We assume the agent possesses a prior probabilistic belief over the pursuer's parameters.
All pursuer parameters are collected into a single vector 
\begin{equation}
\Theta_P=[x_P,y_P,\psi_P,a,R,v_P]^\top,
\end{equation}
which is modeled as Gaussian, $\Theta_P\sim\mathcal{N}(\mu_{\Theta_P},\Sigma_{\Theta_P})$, with mean
\begin{equation}
\mu_{\Theta_P}=[\mu_{x_P},\mu_{y_P},\mu_{\psi_P},\mu_{a},\mu_{R},\mu_{v_P}]^\top, 
\end{equation}
and covariance matrix 
\begin{equation}
    \Sigma_{\Theta_P} = \begin{bmatrix}
        \Sigma_{P} & \mathbf{0}_{2\times1}& \mathbf{0}_{2\times1}&\mathbf{0}_{2\times1}&\mathbf{0}_{2\times1}  \\
        \mathbf{0}_{1\times2} & \sigma_{\psi_P}^2&0 & 0& 0\\
        \mathbf{0}_{1\times2} &0 & \sigma_{a}^2& 0& 0\\
        \mathbf{0}_{1\times2}& 0&0&\sigma_{R}^2& 0\\
        \mathbf{0}_{1\times2}& 0&0&0&\sigma_{v_P}^2
    \end{bmatrix}.
\end{equation}
The mean vector contains the expected values of each pursuer parameter and $\Sigma_{P}$ represents the covariance of the pursuer's position.  The remaining diagonal elements correspond to the variances of the other pursuer parameters. The $x$ and $y$ components of the pursuer's position are assumed correlated, while all other parameters are assumed uncorrelated. If correlations existed among the remaining parameters, the corresponding off-diagonal elements of $\Sigma_{\Theta_P}$ would be nonzero. 
The probability density function (PDF) of $\Theta_P$ is 
\begin{equation}\label{eq:theta_pdf}
    f_{\Theta_P}(\tau)=\frac{1}{\sqrt{(2\pi)^6\,\lvert \Sigma_{\Theta_P}\rvert}}\,
    \exp\!\Biggl(
      -\tfrac{1}{2}\,(\tau - \mu_{\Theta_P})^\top\,\Sigma_{\Theta_P}^{-1}\,(\tau - \mu_{\Theta_P})
    \Biggr).
\end{equation}

The evader’s parameters are similarly combined into a single vector, $\Theta_E = [x_E,y_E,v_E]^\top$.
In this work, the evader’s parameters are treated as known; however, extending the formulation to include uncertainty in $\Theta_E$ would be straightforward and of practical interest.

The CSBEZ expression in Equation~\eqref{eq:ez_equation} can be written as a function of the pursuer and evader parameters, denoted $z(\Theta_P,\Theta_E)$. Given the mean and covariance of the uncertain pursuer parameters $(\mu_{\Theta_P}, \Sigma_{\Theta_P})$, our objective is to compute the probability that the evader lies within the true CSBEZ. We refer to this probability as the CSPEZ. Because the CSPEZ is conditioned on a specific evader configuration $\Theta_E$, its value changes as $\Theta_E$ varies.

The evader is inside the CSBEZ whenever $z(\Theta_P,\Theta_E) \le 0$; therefore, the CSPEZ is the probability that this condition holds under the distribution of $\Theta_P$. Formally, the CSPEZ is obtained by integrating the PDF of $\Theta_P$ over the region where $z(\Theta_P,\Theta_E) \le 0$:
\begin{equation}\label{eq:prob_integral}
    P_{\textit{CSPEZ}}(\mu_{\Theta_P},\Sigma_{\Theta_P},\Theta_E)=P(z(\Theta_P,\Theta_E)\leq0)=\int_{\{\tau|z(\tau,\Theta_E)\leq0\}}f_{\Theta_P}(\tau)d\tau.
\end{equation}
The CSPEZ is the probability the evader lies withing the true CSBEZ, given a specific configuration (evader parameters $\Theta_E$ and pursuer parameter distribution $\mu_{\Theta_P}, \Sigma_{\Theta_P})$. 
This integral lacks a closed-form solution due to the nonlinear coupling among the components of $\Theta_P$ and must therefore be approximated. The integral is evaluated for a specific pursuer distribution -- defined by its mean and covariance -- and a fixed set of evader parameters. Consequently, a new integral must be computed for each unique combination of pursuer distribution and evader configuration. 
For example, if this probability is employed as a constraint within a path-planning algorithm, the integral must be re-evaluated at multiple points along the trajectory using the corresponding evader's position and heading.
In the following sections, we present four methods for approximating this integral: MCCSPEZ, LCSPEZ, QCSPEZ, NNCSPEZ.

\subsection{Monte Carlo Probabilistic Engagement Zone (MCCSPEZ)}
The MCCSPEZ builds upon our previous work~\cite{stagg2025} and uses Monte Carlo integration to evaluate the integral in Equation~\eqref{eq:prob_integral}. 
The MCCSPEZ method begins by drawing $N_m$ random samples from the pursuer-parameter distribution, $\Theta_P^i \sim \mathcal{N}(\mu_{\Theta_P},\Sigma_{\Theta_P}), i = 1,\dots,N_m$. For each random sample, the CSBEZ function is evaluated using the evader parameters, and the number of samples yielding a value less than zero is counted. The MCCSPEZ probability is then approximated as 
\begin{equation}
P_{\textit{MCCSPEZ}}(\mu_{\Theta_P},\Sigma_{\Theta_P},\Theta_E)=
P(z(\Theta_P, \Theta_E) \leq 0) \approx
\frac{1}{N_m}\sum_{i=1}^{N_m}\mathds{1}\left(z(\Theta_P^i,\Theta_E)\right),
\end{equation}
where $\mathds{1}(z)$ is the indicator function defined as
\begin{equation}
    \mathds{1}(z) = \left\{
    \begin{array}{lr}
         1 & z \leq 0\\
         0 & z > 0
    \end{array}\right\}.
\end{equation}
As the number of random samples increases, the accuracy of the MCCSPEZ approximation improves. A drawback of the MCCSPEZ method is the large number of samples, and corresponding computational time, required to achieve an accurate approximation. Another limitation arises when using a gradient-based path-planning algorithm, since computing the Jacobian of the MCCSPEZ is required. However, due to the discontinuous nature of the indicator function, the Jacobian is zero almost everywhere and undefined at the switching boundary.

\subsection{Linearized Probabilistic Engagement Zone (LCSPEZ)}
The LCSPEZ method addresses the large-sample and Jacobian limitation of the MCCSPEZ, albeit at the expense of accuracy.  This method constructs a linear model of the CSPEZ function and uses that linearization to approximate the integral in Equation~\eqref{eq:prob_integral}, yielding a defined solution. A similar approach for the simple BEZ was presented in our prior work~\cite{stagg2025}.

The linear model is obtained using a first-order Taylor-series expansion about the mean of the pursuer parameters.
This is done by computing the Jacobian of the CSPEZ function with respect to the pursuer parameters,
\begin{equation}
    z(\Theta_P+\vect{\delta}_P, \Theta_E ) \approx z(\Theta_P, \Theta_E) + \vect{\delta}_P^\top \frac{\partial z}{\partial \Theta_P},
\end{equation}
where $\partial z / \partial \Theta_P$ is the Jacobian of $z$ with respect to $\Theta_P$.
The Jacobians can be computed either analytically or through automatic differentiation. In this work, we use the JAX automatic-differentiation library~\cite{jax2018github}.
Using this linear model, the mean and variance of the CSBEZ function output can be computed as
\begin{equation}
    \mu_z(\mu_{\Theta_P},\Theta_E)=z(\mu_{\Theta_P},\Theta_E)
\end{equation}
and
\begin{equation}
    \sigma^2_z(\mu_{\Theta_P},\Sigma_{\Theta_P},\Theta_E)=\frac{\partial z}{\partial \Theta_P}\left(\mu_{\Theta_P},\Theta_E\right) \Sigma_{\Theta_P}\frac{\partial z}{\partial \vect{\Theta}_P}\left(\mu_{\Theta_P},\Theta_E\right)^\top.
\end{equation}
We then find the LCSPEZ probability as 
\begin{equation}\label{eq:LCSPEZ}
    P_{\textit{LCSPEZ}}(\mu_{\Theta_P},\Sigma_{\Theta_P},\Theta_E)=P(z(\Theta_P, \Theta_E) \leq 0) \approx F\left(0; \mu_z(\mu_{\Theta_P},\Theta_E),\sigma_z^2(\mu_{\Theta_P},\Sigma_{\Theta_P},\Theta_E)\right),
\end{equation}
where $F(x;\mu,\sigma^2)$ is the single-variable Gaussian cumulative distribution function (CDF). The CDF can be written in terms of the standard error function as
\begin{equation}\label{eq:guassian_cdf}
F(x;\mu,\sigma^2)
= \int_{-\infty}^{x}
    \frac{1}{\sqrt{2\pi}\,\sigma}
    \exp\!\Bigl(-\frac{(t-\mu)^2}{2\sigma^2}\Bigr)
  \,\mathrm{d}t
= \frac{1}{2}\Bigl[1 + \mathrm{erf}\!\bigl(\tfrac{x-\mu}{\sigma\sqrt{2}}\bigr)\Bigr].
\end{equation}

It is important to note that both the mean $\mu_z(\mu_{\Theta_P}, \Theta_E)$ and the variance $\sigma_z^2(\mu_{\Theta_P}, \Theta_E)$ depend on the evader parameters (position, heading, and speed). If the LCSPEZ is used as a constraint within a path-planning algorithm, a distinct mean and variance must be evaluated at points sampled along the trajectory, based on the position and heading defined by the path. The corresponding LCSPEZ values can then be computed at each point using the Gaussian CDF.
In our formulation, we treat the pursuer's parameter distribution as fixed throughout the trajectory. However, if the distribution were time-varying—such as if the covariance evolved over time or the initial position of the pursuer were dynamic—then the relevant parameters would also need to be evaluated as functions of time and sampled along the path.

The LCSPEZ requires less memory and computational complexity than the MCCSPEZ, albeit with reduced accuracy. Specifically, the LCSPEZ performs poorly when the CSBEZ function (Equation~\eqref{eq:ez_equation}) is highly nonlinear, contains discontinuities, or when its Jacobian is near zero. Regions where the Jacobian approaches zero occur at the switching boundaries of the projected pursuer trajectory, for example when the optimal path transitions from a right–straight maneuver to a left–straight maneuver or the reverse transition. Discontinuities in the CSBEZ arise when the final evader position crosses the pursuer's minimum-turn-radius boundary, moving from inside the turn circle to outside, or vice versa. These structural features appear in consistent patterns across the parameter space, which makes the situations where the LCSPEZ breaks down predictable. The LCSPEZ also degrades in performance under high levels of uncertainty.

\subsection{Quadratic Probabilistic Engagement Zone (QCSPEZ)}
Instead of constructing a linear model of the CSBEZ function, the QCSPEZ constructs a quadratic model. This approach is useful when the CSBEZ function exhibits stronger nonlinearity or when its Jacobian is near zero. Rather than a first-order Taylor expansion, a second-order approximation is used:
\begin{equation}
z(\Theta_P + \vect{\delta}_P, \Theta_E)
\approx z(\Theta_P, \Theta_E)
+ \vect{\delta}_P^T
  \frac{\partial z}{\partial \Theta_P}\bigl(\Theta_P,\Theta_E\bigr)
+ \tfrac12\,
  \vect{\delta}_P^T
  \frac{\partial^2 z}{\partial \Theta_P^2}\bigl(\Theta_P,\Theta_E\bigr)
  \,\vect{\delta}_P,
\end{equation}
where $\partial^2 z/\partial \Theta_P^2$ is the Hessian of the CSPEZ function with respect to the pursuer parameters. Unlike the LCSPEZ, a quadratic function of a Gaussian random variable does not admit a closed-form solution and therefore, must be approximated. We do this by first finding the mean value of the quadratic approximation
\begin{equation}
    \mu_z(\mu_{\Theta_P},\Sigma_{\Theta_P},\Theta_E)= z(\mu_{\Theta_P}, \Theta_E)
+ \tfrac{1}{2} \operatorname{tr}\left(
  \frac{\partial^2 z}{\partial \Theta_P^2}(\mu_{\Theta_P}, \Theta_E)\, \Sigma_{\Theta_P}
\right)
\end{equation}
and variance
\begin{equation}
    \sigma_z^2(\mu_{\Theta_P},\Sigma_{\Theta_P}, \Theta_E)=
  \frac{\partial z}{\partial \Theta_P}(\mu_{\Theta_P}, \Theta_E)
\Sigma_{\Theta_P}
  \frac{\partial z}{\partial \Theta_P}(\Theta_P, \Theta_E)
^\top
+ 2\, \operatorname{tr}\left(
    \frac{\partial^2 z}{\partial \Theta_P^2}(\mu_{\Theta_P}, \Theta_E)
  \Sigma_{\Theta_P}
    \frac{\partial^2 z}{\partial \Theta_P^2}(\mu_{\Theta_P}, \Theta_E)
  \Sigma_{\Theta_P}
\right).
\end{equation}
Using this mean and variance we then approximate the QCSPEZ probability as 
\begin{equation}\label{eq:QCSPEZ}
    P_{\textit{QCSPEZ}}(\mu_{\Theta_P},\Sigma_{\Theta_P},\Theta_E)=P(z(\Theta_P, \Theta_E) \leq 0) \approx F(0; \mu_z(\mu_{\Theta_P},\Theta_E),\sigma_z^2(\mu_{\Theta_P},\Sigma_{\Theta_P},\Theta_E)),
\end{equation}
where $F$ is the single-variable Gaussian CDF, defined in Equation~\eqref{eq:guassian_cdf}. Just like the LCSPEZ, there is a unique mean and variance for a given set of evader parameters (position, heading, speed). If the QCSPEZ is used as a constraint within a path-planning algorithm, the mean and variance must be evaluated at sampled points along the trajectory. The QCSPEZ probability can then be found using the Gaussian CDF.

Similar to the LCSPEZ, the QCSPEZ performs poorly when the CSPEZ function is highly nonlinear or discontinuous. However, it provides a better approximation than the LCSPEZ when the gradient is near zero, as it incorporates second-order information from the Hessian.

\subsection{Neural Network Probabilistic Engagement Zone (NNCSPEZ)}
Instead of first approximating the CSBEZ equation, the NNCSPEZ learns a direct mapping from the pursuer’s parameter mean and covariance, together with the evader’s parameters, to the probability that the evader lies within the CSBEZ.
This is achieved using a multilayer perceptron (MLP) regressor that predicts this probability from the combined input features. 

To reduce the input dimensionality and exploit problem symmetries, the MLP input is expressed in a relative coordinate frame: the pursuer is placed at the origin with zero heading, and the evader’s position and heading are expressed relative to this frame.
The combined input vector for the MLP is defined as:
\begin{equation}
x_{\mathrm{NN}}(\mu_{\Theta_P},\Sigma_{\Theta_P},\Theta_E)
=
\begin{bmatrix}
\mu_{a}\\
\mu_{R}\\
\mu_{v_P}\\[6pt]
\sigma_{x_P}^2\\
\sigma_{y_P}^2\\
\sigma_{x_P y_P}\\[6pt]
\sigma_{\psi_P}^2\\
\sigma_{a}^2\\
\sigma_{R}^2\\
\sigma_{v_P}^2\\[6pt]
x_{E}-\mu_{x_P}\\
y_{E}-\mu_{y_P}\\
\psi_{E}-\mu_{\psi_P}\\
v_{E}
\end{bmatrix}
\;\in\;\mathbb{R}^{14}.
\end{equation}
The neural network then learns a function $f_{\phi} \colon \mathbb{R}^{14} \to [0, 1]$, where \( f_{\phi}(x_{\mathrm{NN}}) \) approximates the probability that the evader lies within the true CSBEZ, given the specified uncertainty and relative configuration.

We implement the probabilistic engagement regressor as a fully connected MLP using the Flax framework~\cite{flax2020github}. The network receives as input the 14-dimensional feature vector $x_{\mathrm{NN}} \in \mathbb{R}^{14}$ and propagates it through four hidden layers with widths of 512, 256, 256, and 128 neurons, respectively. Each hidden layer applies layer normalization followed by a sigmoid linear unit (SiLU) activation. Formally, the hidden activations are computed as:
\begin{equation}
\begin{aligned}
h^{(0)} &= x_{\mathrm{NN}}, \\[4pt]
h^{(i)} &= \mathrm{SiLU}\!\left(\mathrm{LayerNorm}\bigl(W^{(i)} h^{(i-1)} + b^{(i)}\bigr)\right),
\quad i = 1, \dots, 4,
\end{aligned}
\end{equation}
where the weight matrices are
\begin{equation}
\begin{gathered}
W^{(1)} \in \mathbb{R}^{512 \times 14}, \quad
W^{(2)} \in \mathbb{R}^{256 \times 512}, \\[4pt]
W^{(3)} \in \mathbb{R}^{256 \times 256}, \quad
W^{(4)} \in \mathbb{R}^{128 \times 256},
\end{gathered}
\end{equation}
with corresponding bias vectors $b^{(i)}$. A final output layer defines the output of the neural network:
\begin{equation}
f_{\phi}(x_{\mathrm{NN}}) = \sigma\bigl(w\,h^{(4)} + b\bigr),
\end{equation}
where $w \in \mathbb{R}^{1 \times 128}$ and $b \in \mathbb{R}$ are the weights and bias of the output layer, and 
\begin{equation}
\sigma(z) = \frac{1}{ (1 + e^{-z})},
\end{equation}
is the sigmoid function, ensuring that $f_{\phi}(x_{\mathrm{NN}}) \in [0, 1]$. The complete set of parameters $\phi$ includes all weight matrices $W^{(i)}$, bias vectors $b^{(i)}$, the output weights $w$, and output bias $b$.

To train the network, we generate $N_{T}$ samples $\{x_{\mathrm{NN}}^{(i)}\}_{i=1}^{N_{T}}$ using Latin hypercube sampling~\cite{mckay2000comparison}, drawing uniformly over the admissible range of each feature. For each sample, we compute the corresponding Monte Carlo CSPEZ probability $P_{\mathit{MCCSPEZ}}^{(i)}$, which serves as the ground truth. We then optimize the network parameters $\phi$ using the Adam optimizer~\cite{kingma2014adam} to minimize the root mean square error (RMSE) loss:
\begin{equation}
\mathcal{L}(\phi)
= \sqrt{\frac{1}{N_{T}}\sum_{i=1}^{N_{T}}\left(f_{\phi}\bigl(x_{\mathrm{NN}}^{(i)}\bigr) - P_{\textit{MCCSPEZ}}^{(i)}\right)^{2}}\,.
\end{equation}

Using the trained network, the NNCSPEZ is defined as
\begin{equation}\label{eq:NNCSPEZ}
    P_{\textit{NNCSPEZ}}(\mu_{\Theta_P}, \Sigma_{\Theta_P}, \Theta_E)
    = f_{\phi}\bigl(x_{\mathrm{NN}}(\mu_{\Theta_P}, \Sigma_{\Theta_P}, \Theta_E)\bigr).
\end{equation}
When trained effectively, the NNCSPEZ achieves higher accuracy than the LCSPEZ and QCSPEZ approximations while requiring significantly less memory than MCCSPEZ. 
An additional advantage is that the Jacobian of the network output with respect to its inputs can be computed analytically via automatic differentiation, enabling seamless integration into gradient-based path-planning algorithms, where gradient information can be directly leveraged during optimization.

\section{Path Planning Using CSPEZ}\label{sec:path}
To illustrate the utility of the CSPEZ formulations, we now present a trajectory optimization framework that incorporates CSPEZ-based constraints. In contested environments, agents must compute safe trajectories that avoid potential engagements while accounting for uncertainty in adversary information. Prior work~\cite{dillon2023optimal,weintraub2022optimal} used cardioid-shaped EZs to compute minimum-time paths that avoid threats. The BEZ formulation in~\cite{von2023basic} extended this concept to model an infinite-turn-rate pursuer for conflict-free trajectory planning. In our earlier work~\cite{stagg2025}, we introduced a linearized probabilistic engagement zone (PEZ) to handle parameter uncertainty in the planning process. Here, we build on that foundation by integrating the CSPEZ approximations—LCSPEZ, QCSPEZ, and NNCSPEZ—directly as constraints within the trajectory optimization routine.

We use B-splines to parameterize our trajectories because they are easily differentiable and have local support (sparse Jacobians) to aid in trajectory optimization. B-splines are parameterized by a set of control points $\mylist{C} = (\vect{c}_1, \vect{c}_2, \hdots{}, \vect{c}_{{N_c}})$, where $N_c$ is the number of control points, and knot points $\vect{t}_k = (t_0-k\Delta_t,\hdots,t_0-\Delta_t,t_0, t_0+\Delta_t, \hdots, t_f, t_f+\Delta_t,\hdots,t_f+k\Delta_t)$, where $t_0$ is the starting time of the trajectory, $t_f$ is the final time of the trajectory, $k$ is the order of the B-spline, $\Delta_t = (t_f-t_0)/N_k$ is the time spacing of the knot points, and $N_k$ is the number of internal knot points. The B-spline path is then defined as a weighted sum of basis functions where the basis functions are defined using the knot points and the weights are the control points
\begin{equation} \label{eq:b-spline_basis}
\vect{p}(t) = \sum ^{{N_c}}_{i=1} B_{i,k}(t)\vect{c}_i.
\end{equation}
The basis functions are defined using the Cox-De-Boor recursive formula~\cite{cox1972numerical}.

The goal of the path-planning algorithm is to find the minimum time trajectory (by optimizing the B-spline control points and knot points) for the evader starting at an initial position $E_0$ and ending at a goal location $E_f$. The CSPEZ approximations are used as constraints to ensure the probability of entering the true CSBEZ is held below a threshold $\epsilon$.

We also employ kinematic feasibility constraints that are defined using differential flatness, assuming the evader follows a unicycle kinematic model~\cite{mellinger2011minimum,tang2010differential}:
\begin{equation} \label{eq:dynamics}
\begin{bmatrix}
\dot{x}_E(t) \\
\dot{y}_E(t) \\
\dot{\psi}_E(t)
\end{bmatrix} = 
\begin{bmatrix}
v_E(t)\cos{\psi_E(t)} \\
v_E(t)\sin{\psi(t)} \\
u_E(t)
\end{bmatrix},
\end{equation}
where it controls its turn rate $u_E(t)$ and travels at constant velocity $v_E.$
Under these kinematics, the velocity of the trajectory is 
\begin{equation}\label{eq:cspez_vel}
  v_E(t) = ||\dot{\vect{p}}(t)||_2,
\end{equation}
where $\vect{\dot{p}}(t)$ is the first time derivative of the B-spline trajectory.
The turn rate is 
\begin{equation}\label{eq:cspez_turn_rate}
    u_E(t) = \frac{\dot{\vect{p}}(t) \times \ddot{\vect{p}}(t)}{||\dot{\vect{p}}(t)||_2^2},
\end{equation}
where $\vect{\ddot{p}}(t)$ is the second time derivative of the B-spline.
Finally, the curvature of the evader's trajectory is 
\begin{equation}\label{eq:cspez_curve}
    \kappa_E(t) = \frac{u(t)}{v(t)}.
\end{equation}

The path optimization problem is 
\begin{subequations} \label{eq:optimization}
\begin{eqnarray}
\mylist{C}_{opt},t_{f_{opt}} = \operatorname*{argmin}_{\mylist{C},t_f}t_f\label{eq:optimization_util}\\
\text{subject to}\
 \vect{p}(0) = E_0 \label{eq:start_constraint}\\
 \vect{p}(t_f) = E_f\label{eq:end_constraint}\\
 \vect{p}(\vect{t}_s) \in \mathcal{D}\label{eq:optimization_path_boundary_constraint}\\
 P_{\textit{CSPEZ}}\bigl(\mu_{\Theta_P},\Sigma_{\Theta_P},\Theta_E(\vect{t}_s)\bigr) \leq \epsilon\label{eq:opt_pez_constraint}\\
v_E(\vect{t}_s) = v_E\label{eq:velocity_constraint}\\
u_{lb} \leq u_E(\vect{t}_s) \leq u_{ub}\label{eq:turn_rate_constraint}\\
-\kappa_{ub} \leq \kappa_E(\vect{t}_s) \leq \kappa_{ub}\label{eq:curve_constraint},
\end{eqnarray}
\end{subequations}
where $u_{lb}$ and $u_{ub}$ are the lower and upper turn-rate constraints, $\kappa_{ub}$ is the path curvature constraint and $u_E(t)$ and $\kappa_E(t)$ denote the turn rate and curvature of the trajectory.
The constraints must be discretely sampled, $\vect{t}_s = \{0, \Delta_s, 2\Delta_s, \hdots, t_f\}$, $\Delta_s = t_f/N_s$, where $N_s$ is the number of discrete constraint samples.
The first two constraints ensure that the trajectory starts and ends at the desired points. The third constraint \eqref{eq:optimization_path_boundary_constraint} ensures that the evader remains within the desired operating region. The next constraint~\eqref{eq:opt_pez_constraint}, is the CSPEZ constraint, which maintains that the probability of the path being within the true CSBEZ is kept below the threshold $\epsilon$. We alternatively use LCSPEZ, QCSPEZ, and NNCSPEZ to approximate this constraint.  In the next section we will show how these three different constraint methods compare. 
The next three constraints ensure that the evader’s trajectory is kinematically feasible. The velocity constraint~\ref{eq:velocity_constraint} limits the trajectory’s speed using Equation~\eqref{eq:cspez_vel}. The turn-rate constraint~\ref{eq:turn_rate_constraint} and the curvature constraint~\ref{eq:curve_constraint}, defined by Equations~\eqref{eq:cspez_turn_rate} and~\eqref{eq:cspez_curve}, ensure that the trajectory remains within the evader’s kinematic limits.

We use a gradient-based interior point optimization algorithm called IPOPT~\cite{ipopt} to perform the optimization problem in Equation~\eqref{eq:optimization}. Because we use a gradient-based optimization algorithm, the Jacobians of all constraints are needed. We employ JAX~\cite{jax2018github} for automatic differentiation. The MCCSPEZ approximation is not used in the path-planning algorithm because its Jacobian is zero almost everywhere due to the flat indicator function, despite being the most accurate approximation.

\section{Results}\label{sec:results}
In this section, we present results demonstrating the utility of the CSPEZ formulation. First, we evaluate the accuracy of the LCSPEZ, QCSPEZ, and NNCSPEZ approximations compared to the MCCSPEZ baseline. Then we show results for the path-planning algorithm when each CSPEZ approximation is used as a constraint. 
\begin{table}[h]
\caption{\label{tab:error} Error metrics for each CSPEZ approximation.}
\centering
\begin{tabular}{lccc}
\hline
Approximation& RMSE& Average Absolute Error&Max Error\\
\hline
LCSPEZ&  0.01343&0.0655 &0.9160\\
QCSPEZ&0.00792 &0.0573 &0.8522\\
NNCSPEZ&\textbf{2.645e-6} &\textbf{0.0009} &\textbf{0.0448}\\
\hline
\end{tabular}
\end{table}

\subsection{CSPEZ Comparison}
To evaluate the accuracy of each approximation, we generate a test set consisting of $N_T = 500{,}000$ samples. Each sample defines a complete CSPEZ configuration consisting of evader parameters, a pursuer mean, and a pursuer covariance. These samples are drawn independently from uniform distributions over the admissible ranges of each parameter. We denote the resulting sets as
\begin{equation}
\begin{aligned}
\{\Theta_E^i\}_{i=1}^{N_T} &= \mylist{\Theta}_E, \\
\{\mu_{\Theta_P}^i\}_{i=1}^{N_T} &= \mylist{\mu}_{\Theta_P}, \\
\{\Sigma_{\Theta_P}^i\}_{i=1}^{N_T} &= \mylist{\Sigma}_{\Theta_P},
\end{aligned}
\end{equation}
where each tuple $(\mu_{\Theta_P}^i, \Sigma_{\Theta_P}^i,\Theta_E^i)$ represents a unique configuration over which the CSPEZ probability approximations are evaluated.
These points are generated separately from the points used to train the NNCSPEZ model. 

To quantify the accuracy of each approximation method, we evaluate the root mean squared error (RMSE), average absolute error (AAE), and maximum absolute error (MaxAE) relative to the MCCSPEZ baseline. These metrics are computed over a test set of $N_T = 500{,}000$ configurations. For each sample $i$, let 
$p_{\text{approx}}^i = P_{\textit{APPROX}}(\mu_{\Theta_P}^i, \Sigma_{\Theta_P}^i, \Theta_E^i)$ 
denote the predicted CSPEZ probability from a given approximation method—specifically, LCSPEZ (Equation~\eqref{eq:LCSPEZ}), QCSPEZ (Equation~\eqref{eq:QCSPEZ}), or NNCSPEZ (Equation~\eqref{eq:NNCSPEZ}). Let 
$p_{\text{MCCSPEZ}}^i = P_{\textit{MCCSPEZ}}(\mu_{\Theta_P}^i, \Sigma_{\Theta_P}^i, \Theta_E^i)$ 
denote the corresponding probability from the MCCSPEZ baseline. The error metrics are then defined as
\begin{equation}
\text{RMSE} = \sqrt{\frac{1}{N_T} \sum_{i=1}^{N_T} \left( p_{\text{approx}}^i - p_{\text{MCCSPEZ}}^i \right)^2},
\end{equation}
for the root mean square error, 
\begin{equation}
\text{AAE} = \frac{1}{N_T} \sum_{i=1}^{N_T} \left| p_{\text{approx}}^i - p_{\text{MCCSPEZ}}^i \right|,
\end{equation}
for the absolute error, and 
\begin{equation}
\text{MaxAE} = \max_{i} \left| p_{\text{approx}}^i - p_{\text{MCCSPEZ}}^i \right|,
\end{equation}
for the maximum error.

RMSE penalizes larger errors more heavily and provides an overall measure of approximation accuracy. AAE represents the average magnitude of the approximation error, while MaxAE captures the worst-case deviation from the baseline. The results of these metrics for each method are shown in Table~\ref{tab:error}.
From this table and figure, we observe that the NNCSPEZ provides the most accurate approximation. However, in many situations the LCSPEZ and QCSPEZ perform adequately. 

\begin{figure}[H]
\centering
\includegraphics[]{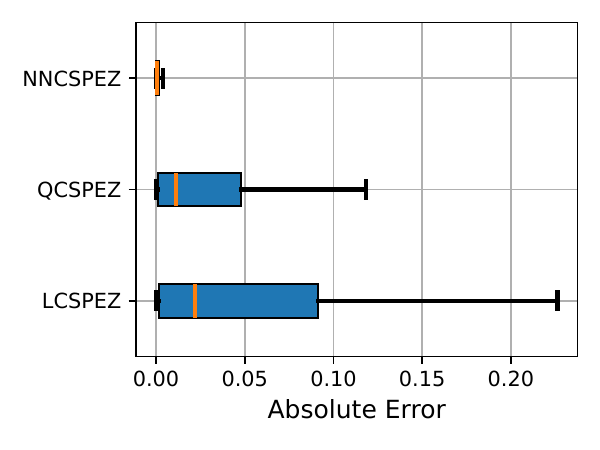}
\caption{Box plot of absolute errors for each CSPEZ approximation relative to the MCCSPEZ baseline; NNCSPEZ shows the lowest error.}\label{fig:error_boxplot}
\end{figure}

In addition to the summary statistics, Figure~\ref{fig:error_boxplot} shows the full distribution of absolute errors across all test points, providing insight into the spread and variability of errors across methods. The NNCSPEZ performs best overall, although both LCSPEZ and QCSPEZ achieve reasonable accuracy across a wide range of configurations.The boxplot shows the median (orange), the 25\% and 75\% percentiles (blue), and the full range of the error (black). As shown, many configurations yield low error for both the QCSPEZ and LCSPEZ. The LCSPEZ has an absolute error below $0.02$ in more than 50\% of the configurations, while the QCSPEZ has an absolute error below $0.01$ in more than 50\% of the configurations. This indicates that in many scenarios, the LCSPEZ and QCSPEZ provide a useful approximation to the true CSPEZ.

\begin{figure}[H]
\centering
\includegraphics[]{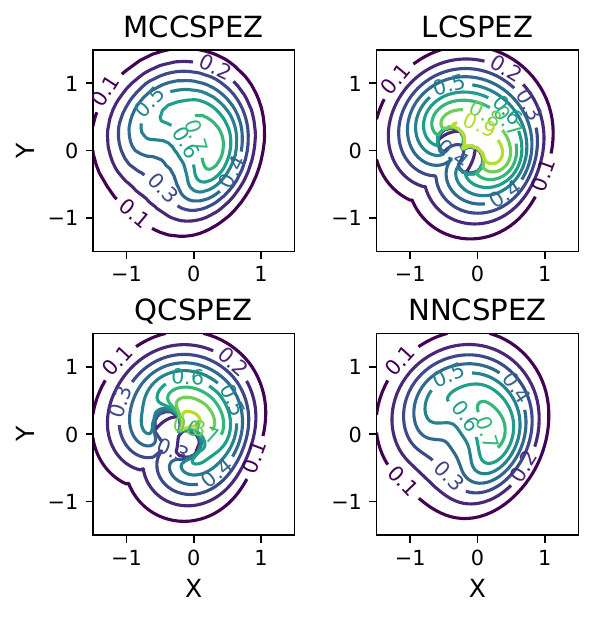}
\caption{Probability level sets for an example CSPEZ scenario. The pursuer’s mean heading is $\pi/2$ (red marker). NNCSPEZ most closely matches the MCCSPEZ baseline.}\label{fig:pez_example}
\end{figure}

To illustrate where the LCSPEZ and QCSPEZ perform well, we show an example scenario where the pursuer's parameter mean value is  $\mu_{\Theta_P}=[0.0,0.0,\pi/4,0.2,1.0,2.0]^\top$ and covariance is 
\begin{equation}
    \Sigma_{\Theta_P} = \begin{bmatrix}
        \begin{bmatrix}
           0.025&0.04\\
           0.04&0.1
        \end{bmatrix} & \mathbf{0}_{2\times1}& \mathbf{0}_{2\times1}&\mathbf{0}_{2\times1}&\mathbf{0}_{2\times1}  \\
        \mathbf{0}_{1\times2} & 0.2&0 & 0& 0\\
        \mathbf{0}_{1\times2} &0 & 0.005& 0& 0\\
        \mathbf{0}_{1\times2}& 0&0&0.1& 0\\
        \mathbf{0}_{1\times2}& 0&0&0&0.3
    \end{bmatrix}.
\end{equation}
We fix the evader heading at $\psi_E=0.0$ and evaluate each model on a uniform grid of evader positions in the $x$-$y$ plane. At each point, the CSPEZ probability is computed using the given approximation method, while holding the pursuer parameters constant. The resulting probability values are visualized using level sets, where each contour corresponds to a fixed CSPEZ probability threshold. The level sets represent contours of constant probability of lying within the true CSBEZ. For example, if the evader remains outside the $0.1$ contour, then the probability of being inside the true CSBEZ is less than $10\%$.
These level sets are shown in Figure~\ref{fig:pez_example}. As shown, the overall shape of the outer level sets ($0.1$–$0.3$) produced by all methods resembles that of the MCCSPEZ. However, both the QCSPEZ and LCSPEZ struggle to accurately capture the larger value level sets.

\begin{figure}[H]
\centering
\includegraphics[]{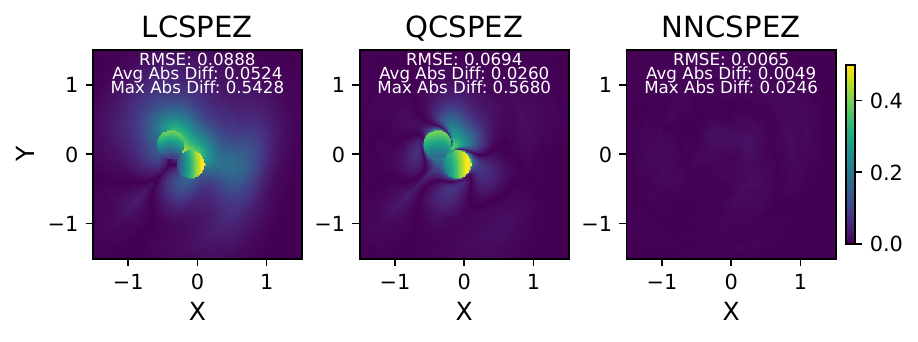}
\caption{Absolute error between each approximation and the MCCSPEZ baseline for the example scenario. The pursuer’s mean heading is $\pi/2$; RMSE, mean, and max errors are shown in each plot.}\label{fig:pez_example_error}
\end{figure}

To assess the accuracy of each approximation, we compute the absolute error between the predicted probability and the MCCSPEZ baseline at each grid point. These spatial error distributions are shown in Figure~\ref{fig:pez_example_error}.
In this scenario, the mean value of the pursuer's heading points in the positive $y$ direction. The largest errors for the LCSPEZ and QCSPEZ approximations occur in regions near the shifted turn radii of the pursuer. This corresponds to the cavity region in the deterministic CSBEZ shown in Figure~\ref{fig:CSBEZ}. The high errors in these regions arise from discontinuities in the CSBEZ surface, which reduce the accuracy of the linear and quadratic  approximations. However, the overall shapes of the probability contours remains similar across all methods.

\begin{figure}[H]
\centering
\includegraphics[]{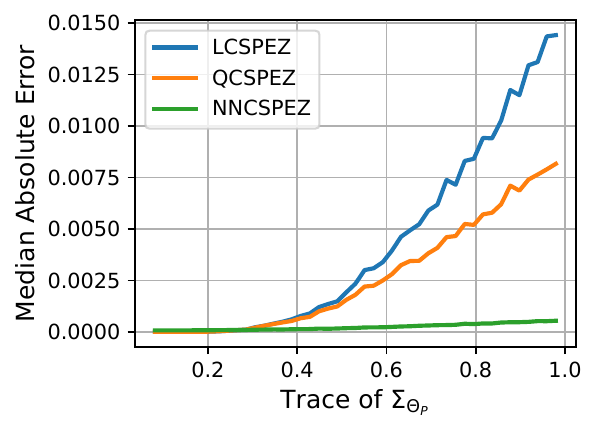}
\caption{Absolute error versus the trace of the pursuer covariance. As uncertainty increases, LCSPEZ and QCSPEZ degrade while NNCSPEZ remains accurate.}\label{fig:error_trace}
\end{figure}

We also show how well each approximation performs under varying levels of uncertainty. To do this, we compute the trace of each pursuer covariance matrix to represent the overall level of uncertainty for that configuration. The $500{,}000$ test samples are sorted by trace and grouped into uniformly spaced bins. Within each bin, we compute the median of the absolute error between each approximation and the MCCSPEZ baseline. This analysis is shown in Figure~\ref{fig:error_trace}.

As the level of uncertainty increases, the accuracy of both the LCSPEZ and QCSPEZ deteriorates, while the NNCSPEZ remains stable. This is expected, since both the linear and quadratic approximations are local to the mean value, and higher uncertainty leads to larger deviations from the region where the approximation is valid. In contrast, the NNCSPEZ approximation does not exhibit a strong dependence on uncertainty, as it is trained directly on CSPEZ probability values. However, the NNCSPEZ can suffer in configurations that are far from its training distribution.

\subsection{Path Planning}
To evaluate the impact of each approximation on trajectory optimization, we use the CSPEZ probability as a constraint within the path-planning framework described in Section~\ref{sec:path}. At each discretized point along the B-spline trajectory, the CSPEZ probability is evaluated using the selected approximation method, and a constraint is enforced to ensure that the probability of the evader entering the engagement zone remains below a user-specified threshold.

The same mean and covariance of the pursuer parameters as in Figure~\ref{fig:pez_example} are used for this test. The evader’s speed is $v_E = 1.0$, with turn-rate and curvature limits of $\pm1.0~\text{rad/s}$ and $0.2~\text{rad/m}$, respectively. The evader starts at $(-4.0,-4.0)$ and must reach $(4.0,4.0)$. The B-spline trajectory has $N_c = 8$ control points and degree $k = 3$. CSPEZ probability thresholds of $[0.01, 0.05, 0.25, 0.5]$ are tested.

\begin{figure}[h]
\centering
\includegraphics[]{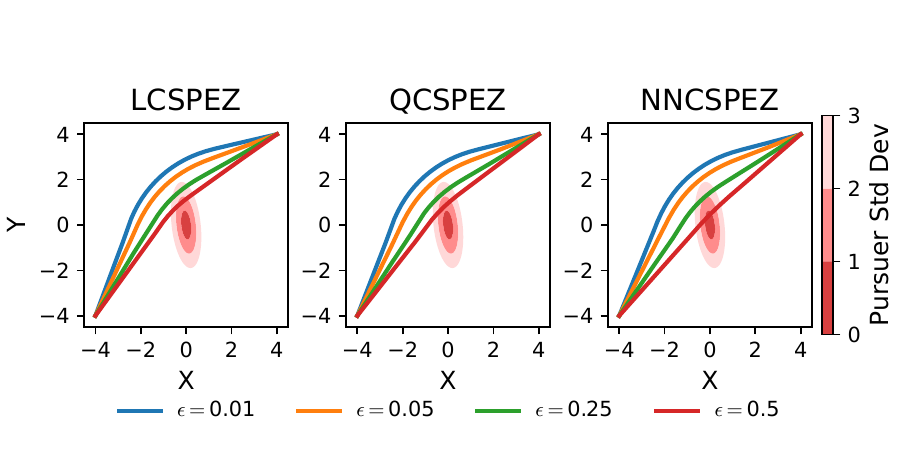}
\caption{Planned trajectories using LCSPEZ (left), QCSPEZ (middle), and NNCSPEZ (right) as CSPEZ constraints. Constraint levels $[0.01, 0.05, 0.25, 0.5]$ are shown, with Mahalanobis distances on the pursuer's position of one, two, and three shown in gradations of red.}\label{fig:path_planning_comparison}
\end{figure}

Figure~\ref{fig:path_planning_comparison} shows the resulting trajectories when using the LCSPEZ, QCSPEZ, and NNCSPEZ approximations as CSPEZ constraints.
For each case, we report three quantities: (1) the final time $t_f$ (proportional to path length), (2) the maximum MCCSPEZ probability along the trajectory (the least safe point), and (3) the total optimization time.
These results are summarized in Table~\ref{tab:path_planning}. NNCSPEZ consistently yields trajectories that most closely match the MCCSPEZ baseline. LCSPEZ and QCSPEZ tend to produce more conservative paths, as indicated by lower maximum MCCSPEZ values. For looser constraints, LCSPEZ often converges faster than NNCSPEZ.

\begin{table}
\caption{\label{tab:path_planning} Path Planning Comparison}
\centering
\begin{tabular}{lccc@{}p{6mm}@{}ccc@{}p{6mm}@{}ccc}
\hline\hline
 & \multicolumn{3}{c}{LCSPEZ} & &
   \multicolumn{3}{c}{QCSPEZ} & &
   \multicolumn{3}{c}{NNCSPEZ} \\
   \cline{2-4}\cline{6-8}\cline{10-12}
$\epsilon$ & $t_f$ & MCCSPEZ & Opt Time & &
           $t_f$ & MCCSPEZ & Opt Time & &
           $t_f$ & MCCSPEZ & Opt Time \\
\cline{2-4}\cline{6-8}\cline{10-12}
0.01 & 11.76 & 0.0013 & 0.5006 && 12.21 & 0.0000 & 0.5943 && 11.62 & 0.0087 & 0.4987 \\
0.05 & 11.58 & 0.0213 & 0.4057 && 11.81 & 0.0010 & 0.5351 && 11.49 & 0.0563 & 0.4277 \\
0.25 & 11.38 & 0.1761 & 0.3453 && 11.42 & 0.1117 & 0.3380 && 11.34 & 0.2562 & 0.4326 \\
0.5  & 11.29 & 0.3973 & 0.2762 && 11.24 & 0.5336 & 0.4150 && 11.25 & 0.4986 & 0.4063 \\
\hline\hline
\end{tabular}
\end{table}

To evaluate the benefit of probabilistic versus deterministic constraints, we also compare against the CSBEZ formulation. The CSBEZ assumes exact pursuer parameters (i.e., no uncertainty), whereas the CSPEZ formulations explicitly incorporate uncertainty, enabling a tunable safety threshold.
A CSPEZ constraint of $0.5$ using LCSPEZ corresponds approximately to enforcing $z(\mu_{\Theta_P}, \Theta_E) \leq 0$ (Equation~\eqref{eq:ez_equation}), equivalent to the deterministic CSBEZ boundary.
As shown in Table~\ref{tab:path_planning}, the probabilistic CSPEZ constraints yield significantly safer paths while maintaining comparable path lengths. This demonstrates that incorporating uncertainty through probabilistic constraints can enhance safety with only modest increases in computational cost and trajectory duration.

\section{Conclusion}\label{sec:conclusion}
This work introduced a framework for modeling probabilistic engagement zones to enable path planning in adversarial environments under uncertainty. The first contribution presented an analytic solution for the CSBEZ. Next, the CSPEZ was presented, which estimates the probability an evader lies within the true CSBEZ when the pursuer’s parameters are uncertain.

Four approximation methods were developed to estimate the CSPEZ probability: a Monte Carlo baseline (MCCSPEZ), a linearized model (LCSPEZ), a quadratic model (QCSPEZ), and a neural network model (NNCSPEZ). MCCSPEZ provides high accuracy but is computationally expensive and non-differentiable. LCSPEZ and QCSPEZ offer analytic and differentiable approximations requiring no training and converge faster under certain conditions. NNCSPEZ achieves the highest accuracy but relies on representative training data near the evaluated configuration.

Each method was demonstrated as a constraint in trajectory optimization, enabling planners to generate paths that explicitly incorporate risk. This enables decision-making in contested regions where deterministic approaches may be unsafe. Results indicated that NNCSPEZ produced the shortest and safest trajectories, while LCSPEZ and QCSPEZ remain practical when training data are unavailable.

This study focused on two-dimensional scenarios to maintain tractable analysis, training, and visualization. Nevertheless, the proposed framework readily extends to three-dimensional vehicle dynamics and more realistic engagement models. Future work will consider uncertainty in the evader state, incorporate multiple threats, and develop trajectory-initialization strategies to enhance convergence.

\section*{Acknowledgments}
This work was supported by the NSF IUCRC Phase I: Center for Autonomous Air Mobility and Sensing (CAAMS) under Award No.~2139551.

Portions of this manuscript were refined and edited using OpenAI’s ChatGPT model (GPT-5). The tool was employed for language polishing, LaTeX and Python code assistance, and to help organize and clarify research ideas. All technical content, algorithms, and conclusions were developed, verified, and approved by the authors.

\bibliography{sample}

\end{document}